\documentclass[12pt,journal,compsoc,onecolumn,draftclsnofoot]{IEEEtran}

\usepackage{scalerel}
\usepackage{tikz}
\usetikzlibrary{svg.path}

\definecolor{orcidlogocol}{HTML}{A6CE39}
\tikzset{
 orcidlogo/.pic={
 \fill[orcidlogocol] svg{M256,128c0,70.7-57.3,128-128,128C57.3,256,0,198.7,0,128C0,57.3,57.3,0,128,0C198.7,0,256,57.3,256,128z};
 \fill[white] svg{M86.3,186.2H70.9V79.1h15.4v48.4V186.2z}
 svg{M108.9,79.1h41.6c39.6,0,57,28.3,57,53.6c0,27.5-21.5,53.6-56.8,53.6h-41.8V79.1z M124.3,172.4h24.5c34.9,0,42.9-26.5,42.9-39.7c0-21.5-13.7-39.7-43.7-39.7h-23.7V172.4z}
 svg{M88.7,56.8c0,5.5-4.5,10.1-10.1,10.1c-5.6,0-10.1-4.6-10.1-10.1c0-5.6,4.5-10.1,10.1-10.1C84.2,46.7,88.7,51.3,88.7,56.8z};
 }
}

\newcommand\orcidicon[1]{\href{https://orcid.org/#1}{\mbox{\scalerel*{
\begin{tikzpicture}[yscale=-1,transform shape]
\pic{orcidlogo};
\end{tikzpicture}
}{|}}}}

\usepackage[T1]{fontenc}
\usepackage[retain-explicit-plus]{siunitx}
\usepackage{xcolor}
\usepackage{booktabs}
\usepackage{tabu}
\usepackage{tabularx}
\usepackage{tabulary}
\usepackage{multirow,bigdelim}
\usepackage{rotating}
\usepackage{makecell}
\usepackage{tablefootnote}
\usepackage{threeparttable}
\usepackage[ruled,vlined]{algorithm2e}
\usepackage{listings}
\usepackage{enumitem}
\usepackage[hyphens]{url}
\usepackage{lineno,hyperref}
\usepackage{comment}
\usepackage{tablefootnote}
\usepackage{enumitem,amssymb}
\usepackage{subcaption}
\usepackage{graphicx}
\usepackage{amssymb}
\usepackage{pifont}

\newlist{todolist}{itemize}{2}
\setlist[todolist]{label=$\square$}

\usepackage{blindtext}
\usepackage{hyperref}
\usepackage{nameref}

\newcounter{mylabelcounter}

\makeatletter
\newcommand{\labelText}[2]{%
\refstepcounter{mylabelcounter}%
\immediate\write\@auxout{%
 \string\newlabel{#2}{{\unexpanded{#1}}{\thepage}{{\unexpanded{#1}}}{mylabelcounter.\number\value{mylabelcounter}}{}}%
}
}
\makeatother

\makeatletter
\newcommand\footnoteref[1]{\protected@xdef\@thefnmark{\ref{#1}}\@footnotemark}
\makeatother

\begin{document}

\textcopyright 2024 IEEE. Personal use of this material is permitted. Permission from IEEE must be obtained for all other uses, in any current or future media, including reprinting/republishing this material for advertising or promotional purposes, creating new collective works, for resale or redistribution to servers or lists, or reuse of any copyrighted component of this work in other works. DOI: \url{https://doi.org/10.1109/MIE.2023.3328106}
\newpage

\title{Informatics \& dairy industry coalition: \textsc{ai} trends and present challenges}

\author{Silvia García-Méndez\orcidicon{0000-0003-0533-1303}, Francisco de Arriba-P\'erez\orcidicon{0000-0002-1140-679X}, María del Carmen Somoza-López\orcidicon{0000-0001-8709-684X}
\IEEEcompsocitemizethanks{\IEEEcompsocthanksitem Silvia García-Méndez, and Francisco de Arriba-Pérez are with the Information Technologies Group, atlanTTic, University of Vigo, Spain. \IEEEcompsocthanksitem María del Carmen Somoza-López is with the Applied Mathematics I Department, University of Vigo, Spain. 
\\E-mail: farriba@gti.uvigo.es}}

\markboth{IEEE Industrial Electronics Magazine}{}
\markboth{IEEE Industrial Electronics Magazine}
{de Arriba-Pérez \MakeLowercase{\textit{et al.}}: Informatics \& dairy industry coalition: \textsc{ai} trends and present challenges}

\makeatletter
\long\def\@IEEEtitleabstractindextextbox#1{\parbox{0.922\textwidth}{#1}}
\makeatother

\IEEEtitleabstractindextext{%
\begin{abstract}
Artificial Intelligence (\textsc{ai}) can potentially transform the industry, enhancing the production process and minimizing manual, repetitive tasks. Accordingly, the synergy between high-performance computing and powerful mathematical models enables the application of sophisticated data analysis procedures like Machine Learning. However, challenges exist regarding effective, efficient, and flexible processing to generate valuable knowledge. Consequently, this work comprehensively describes industrial challenges where \textsc{ai} can be exploited, focusing on the dairy industry. The conclusions presented can help researchers apply novel approaches for cattle monitoring and farmers by proposing advanced technological solutions to their needs.
\end{abstract}

}

\maketitle

\IEEEdisplaynontitleabstractindextext

\IEEEpeerreviewmaketitle

\section{\textsc{ai} at the service of new industrial challenges}

Artificial Intelligence (\textsc{ai}) was created to assist human operators by mimicking human reasoning, often enhancing their performance. In this line, \textsc{ai}-based solutions have the potential to transform the industrial sectors by improving the production process and work environment and minimizing at the same time the manual and repetitive tasks performed by human operators. Consequently, its popularity in real industrial problems is rising \cite{DeVries2023}. 

The latter trend is motivated by the spread of the Internet of Things (\textsc{i}o\textsc{t}) in our everyday lives and the industry due to the affordable and novel hardware and software solutions like big data, cloud computing and the distributed application paradigm for data management and predictive model design. Consequently, a vast amount of data is continuously generated by multiple heterogeneous and intelligent connected devices (\textit{e.g.}, home automation gadgets, machinery, living organisms like cattle, etc.) with exponential growth \cite{Oussous2018}. Mainly, the Industrial Internet of Things (\textsc{ii}o\textsc{t}), one of the pillars of Industry 4.0, consists of the direct exchange of data between communication nodes, that is, among intelligent objects, people, and processes \cite{Jasperneite2020}. More in detail, \textsc{ai} technologies, \textsc{i}o\textsc{t} platforms, and edge computing capabilities created an industrial revolution by enabling not only monitoring but traceability of the value chain towards process optimization and quality guarantee.

The synergy between high-performance computing systems and powerful mathematical models can enable the application of sophisticated procedures for data analysis (\textit{e.g.}, pattern inference, classification, etc.) like Machine Learning (\textsc{ml}) from different existing approaches (\textit{i.e.}, supervised, unsupervised and Deep Learning \cite{DeSilva2020}). However, there are still challenges regarding effective, efficient, and flexible processing of this information to generate valuable insights \cite{Jun2011}, especially in industrial use cases. The ultimate objective is to design appropriate and optimized approaches to solve indexing and scalability problems, running away from costly solutions.

Accordingly, this work contributes with a comprehensive description of industrial challenges where \textsc{ai} can be exploited in the particular use case of the dairy industry (\textit{e.g.}, cattle lameness prediction that affects the loco-motor system and is associated with stress \cite{Haladjian2018}, estrus disorder detection which refers to the period in which a cow is sexually receptive and directly affects reproduction \cite{Heo2019}, etc.). More in detail, representative solutions based on the application of \textsc{ai} are presented, elaborating on their strengths and weaknesses and discussing how these solutions address the sector's pressing concerns regarding effectiveness and efficiency. The data gathered in this work, and the conclusions presented can help researchers apply novel approaches for cattle monitoring and farmers by proposing advanced technological solutions to address their needs. This work also aims to promote coordination among the key players of the dairy industry (\textit{i.e.}, private and public partners and farmers) by providing an overview of the current knowledge in dairy cattle, particularly investigating the opportunities and limitations of precision dairy farming.

The rest of this paper is organized as follows. Section \ref{sec:industrial_informatics} introduces the field of industrial informatics, paying particular attention to the dairy industry (Section \ref{sec:dairy_industry}). Moreover, Section \ref{sec:ml_solutions} discusses the most recent \textsc{ml} contributions to the dairy sector, and Section \ref{sec:challenges} identifies the main challenges. Finally, Section \ref{sec:conclusions} synthesizes our findings, draws some conclusions, and speculates about future research on new \textsc{ai} applications in the field.

\section{Introduction to industrial informatics}
\label{sec:industrial_informatics}

Industrial informatics is a relatively new field comprising data from various sources (\textit{e.g.}, machinery, the workforce, etc.). Industrial enterprises have recently started to acknowledge the great potential of the field, specifically regarding \textsc{ai}-based solutions towards productivity enhancement through automation and the compliance of industry and government policies \cite{Knights2021}. It is the case of CattleEye\footnote{Available at \url{https://www.v7labs.com/case-study/cattleeye}, October 2023.}, a relevant company in the \textit{smart} agriculture sector, whose \textsc{v}7 software is used for cattle recognition and animal annotation, among other capabilities.

In this line, data are extracted using portable devices like wearables and distributed sensors, and then these data are mined to gather valuable insights to promote industrial productivity \cite{Patel2022}. Note that mobile assistants integrated within the production environments also provide valuable information \cite{Tao2019}.

Notably, the bulk of \textsc{ai} solutions nowadays is \textsc{ml}. The latter models exploit analytical and computational techniques to infer the behavior of the target feature given the available data and have been applied to a variety of industrial challenges related to predictive maintenance \cite{gupta2023predictive}, and efficiency analysis \cite{Bhattacharya2020}, among other uses cases.

When it comes to supervised models, they are trained over past data that was previously annotated by experts in order to predict new future events \cite{DeArriba-Perez2020}. These models can be trained either in batch or online \cite{DeSilva2020}. In the first scenario, periodical training of the models is performed with accumulated data, while in the second, the models are updated with the incoming events. The selection of the training approach depends enormously on the specific classification problem to solve \cite{Bisong2019}. Conversely, unsupervised models do not rely on annotated data \cite{DeSilva2020}. Instead, they exploit \textit{ad hoc} functions to discover hidden patterns from the incoming data. Note that an unsupervised learning stage is often included in the \textsc{ml} pipeline of the supervised approach as preprocessing. Finally, Deep Learning models can automatically generate knowledge and are especially useful in complex use cases \cite{Kiranyaz2020}. However, the computational and time costs related to the convergence of the latter models prevent their application in specific real-world problems. Finally, a popular trend is eXplainable \textsc{ai} (\textsc{xai}) to ensure that the predictions of \textsc{ml} models can be interpreted by operators, which is essential to promote trust in \textsc{ai} solutions towards responsible use of its potential \cite{neethirajan2023artificial}.

\section{\textit{Milking} the data in the dairy industry}
\label{sec:dairy_industry}

The new economic and customer demands and technological advancements have caused profound changes in the dairy sector in the last few years. Innovative practices are essential to meet the demands of animal-derived products, which are expected to rise by \SI{70}{\percent} globally in the following decades \cite{Rojas-Downing2017}.

Moreover, dairy farming plays a vital role in the worldwide agriculture industry towards promoting the sustainability of rural areas \cite{Rozhkova2020}. In this line, dairy products represent \SI{14}{\percent} of the global agricultural trade\footnote{\label{responsiblebusiness}Available at \url{https://www.responsiblebusiness.com/news/americas-news/five-ways-technology-is-transforming-the-dairy-industry}, October 2023.}. Similarly, milk production is expected to rise by 177 million tonnes worldwide by 2025\footnoteref{responsiblebusiness}.

\begin{figure*}[!htbp]
\centering
\includegraphics[scale=0.4]{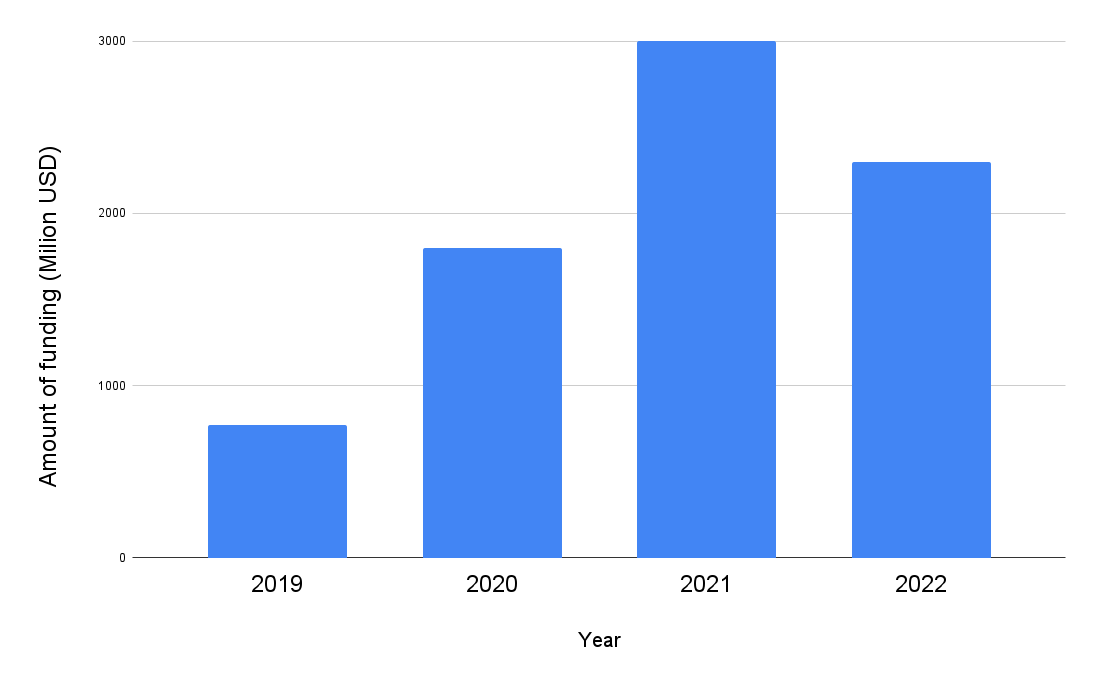}
\caption{\label{fig:funding_chart}Funding trend on precision dairy farming.}
\end{figure*}

\begin{figure*}[!htbp]
\centering
\includegraphics[scale=0.4]{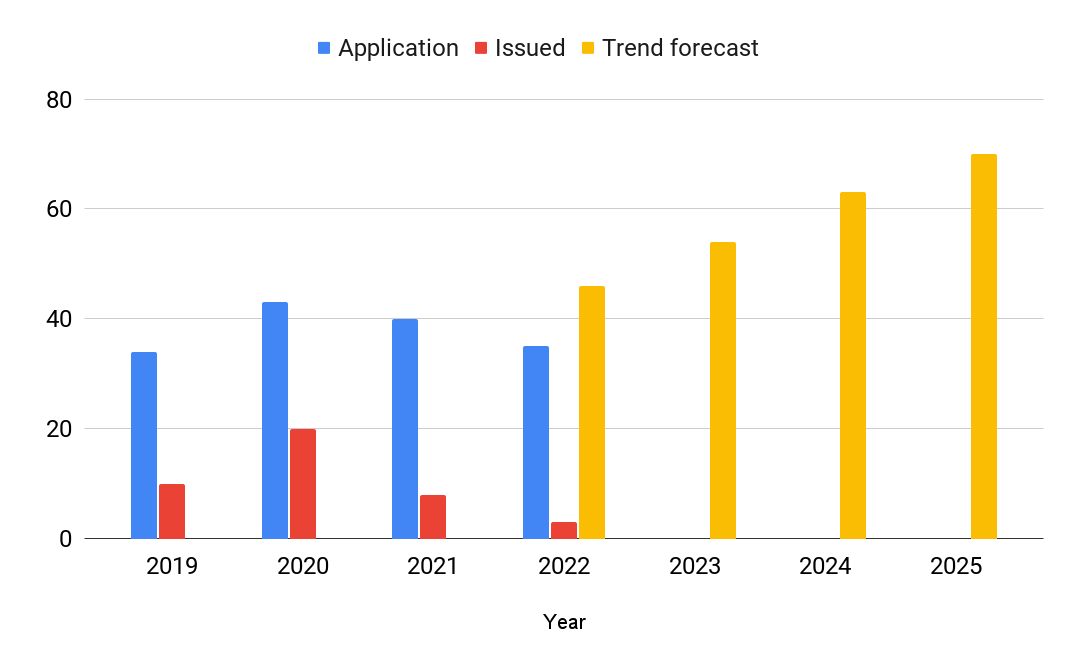}
\caption{\label{fig:patent_chart}Use trend of \textsc{ai} technologies (patent applications, issue rates, trend forecast).}
\end{figure*}

More in detail, in the survey of the dairy industry by McKinsey in 2020\footnote{\label{mckinsey}Available at \url{https://www.mckinsey.com/industries/agriculture/our-insights/whats-ahead-for-the-dairy-industry}, October 2023.}, \SI{80}{\percent} of the executives planned to deploy new digital and analytics technologies in the following years. Even though the executives have recognized the relevance and appropriateness of these technologies, they lack confidence in their current capabilities. Accordingly, \SI{2}{\percent} of the executives surveyed reported using these technologies, while \SI{16}{\percent} claimed to be exploiting large data gathering and insight generation. Figure \ref{fig:funding_chart} shows the rising funding trend on precision dairy farming, while Figure \ref{fig:patent_chart} depicts the use trend of \textsc{ai} technologies in terms of patent applications and issue rates.

Accordingly, farmers are shifting from a traditional to a more innovative approach using novel practices to enhance production \cite{agrawalprecision}. Consequently, applying \textsc{ai} techniques in farming is expected to be extensive in the coming years due to their advantages in terms of expense reduction and production enhancement \cite{Mendes2021}. In this line, improved decision-making through \textsc{ai} has become essential \cite{Lovarelli2020} moved by the population growth trend, which also motivated the growth of herd sizes and derives in the aforementioned higher need for food supplies. Note that by 2025, food demand is expected to leap by \SI{35}{\percent} - \SI{56}{\percent}\footnote{\label{v7labs}Available at \url{https://www.v7labs.com/blog/ai-in-agriculture}, October 2023.}. Innovative technology like \textsc{ai}, robots, and sensors can contribute to fulfilling the food demands of our society \cite{Dzermeikaite2023}.

\begin{figure*}[!htbp]
\centering
\includegraphics[scale=0.19]{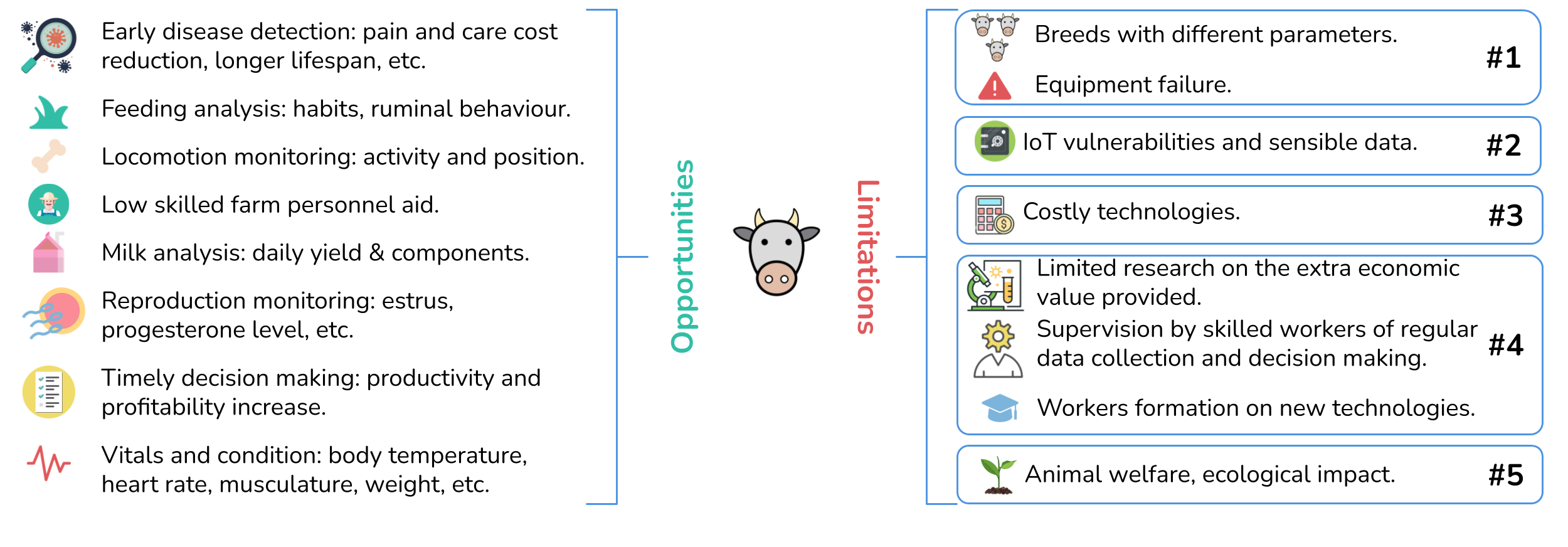}
\caption{\label{fig:scheme}Opportunities and limitations of precision dairy farming.}
\end{figure*}

Precision agriculture, in general, and precision dairy farming, in particular, enable enhanced individual animal management. Precision dairy farming lies on the concept of \textit{per animal approach} towards high quality, quantity, and sustainable production \cite{agrawalprecision} (see Figure \ref{fig:scheme}). Note that the productivity increase per animal is imperative in our scenario of limited resources. Accordingly, by 2025, investment on precision agriculture (\textit{i.e.}, \textsc{ai}- and \textsc{ml}-based solutions) will triple to \$15.3 billion according to a Forbes report\footnote{Available at \url{https://www.forbes.com/sites/louiscolumbus/2021/02/17/10-ways-ai-has-the-potential-to-improve-agriculture-in-2021/?sh=67d79daf7f3b}, October 2023.}. The limitations are classified in terms of the five most relevant challenges of the existing models: compatibility and interoperability (\#1), data privacy \& security (\#2), implementation costs (\#3), specialized training (\#4); and application of environmental-responsible and sustainable practices (\#5) (see discussion in Section \ref{sec:challenges}).

The most common precision dairy farming devices are those for automatic feeding and milk analysis, electronic identification systems, and pedometers. More in detail, intelligent farming equipment compared to devices without data-sharing functionalities \cite{Dzermeikaite2023} also allows environmental management, \textit{i.e.}, measuring the farming impact through their associated emissions \cite{Lovarelli2020}.

\begin{figure*}[!htbp]
\centering
\includegraphics[scale=0.15]{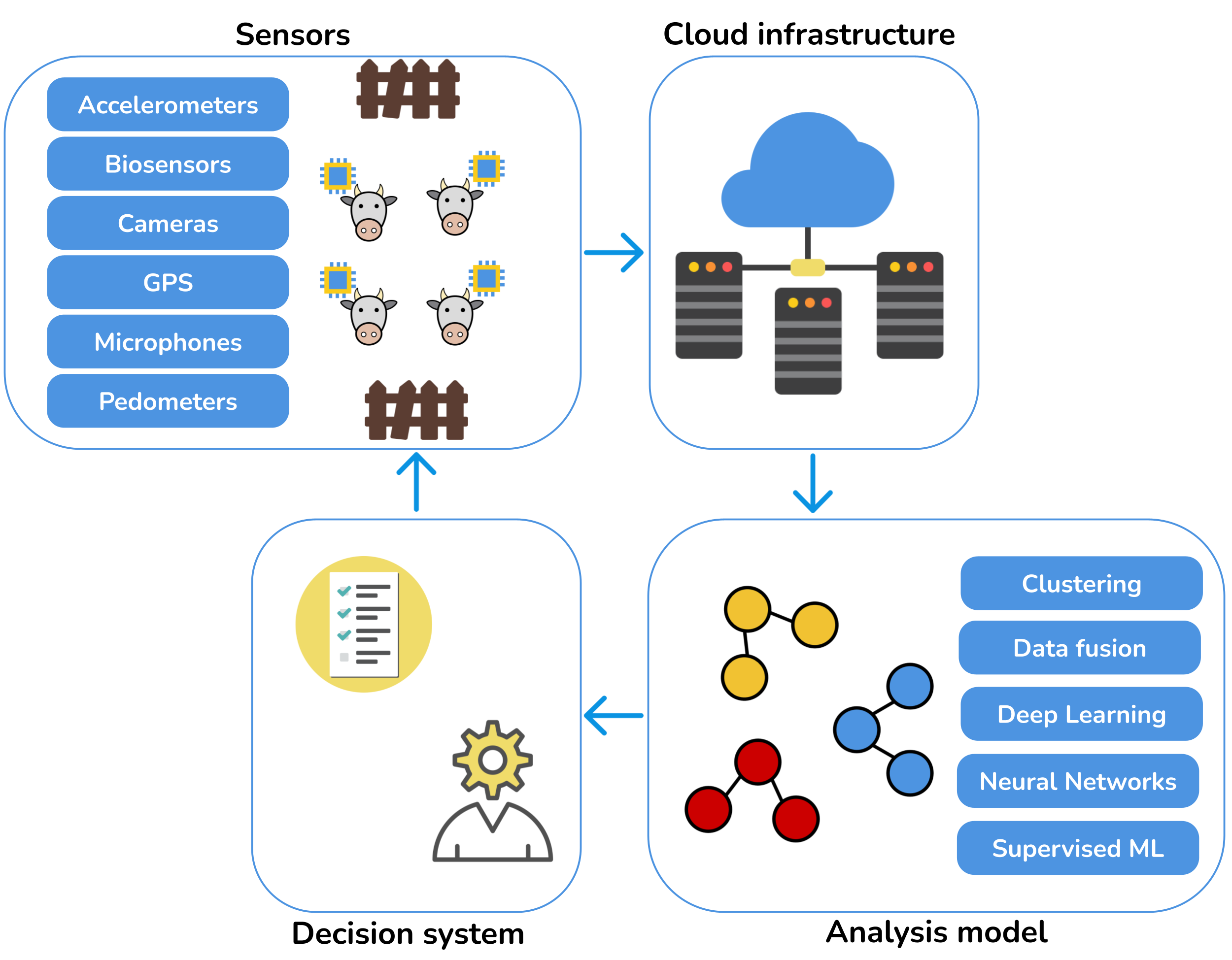}
\caption{\label{fig:precision_farming}Elements of a precision farming solution.}
\end{figure*}

Precision farming is usually performed with the following components: a sensor for data generation, the cloud infrastructure, the analysis model, and finally, a management and implementation decision system, as shown in Figure \ref{fig:precision_farming}. Note that a list of the most popular sensors and representative technologies is included.

Of particular interest is the analysis of image, sound, and video automatically, continuously, and in real-time from data gathered by those above wearable (invasive) and non-wearable (non-invasive), remote sensors like infrared thermal cameras and microphones, and biosensors (\textit{i.e.}, those that measure behavioral, immunological and physiological conditions) \cite{Neethirajan2017}. Note that the latter biosensors have recently become very popular in advanced farming \cite{Dzermeikaite2023}.

The latest technical findings enable intelligent animal monitoring with various devices and techniques that can be transferred to other sectors apart from dairy farming \cite{Dzermeikaite2023}. More in detail, animal behavior, heat, diseases (\textit{e.g.}, lameness, mastitis), and location can be tracked with accelerometers, \textsc{gps} sensors, and pedometers through active time and prolonged down/up time. Pregnancy is detected by milking robots using accelerometers, pedometers, and other sensors to gather activity, milk, and temperature data. Similarly, calving can be analyzed with accelerometers, infrared thermometry, and video processing, taking advantage of \textsc{ai} and \textsc{ml}, intra-vaginal thermometers, and pedometers to gather data about activity and behavior (\textit{e.g.}, up/down and rumination time), temperature, etc. Heat can be studied with accelerometers and pedometers, acoustic sensors, chemical analysis, video processing, and spectroscopy software to extract data related to activity (\textit{e.g.}, rumination), friction, milk, odor, etc. Regarding relevant diseases, lameness can be inferred with accelerometers and pedometers and video processing software using data related to activity and behavior (\textit{i.e.}, eating, laying, rumination, walking). At the same time, mastitis can be detected through accelerometers and pedometers, electrical conductivity and somatic cell count sensors as well as biosensors, image processing, and spectroscopy software to measure behavior (\textit{i.e.}, eating, lying, rumination), milk, and temperature data.

Given the popularity of accelerometers used with cattle that are nowadays commercially available, consideration should be given to those created by Agritech\footnote{Available at \url{https://agritech.com.ve}, October 2023.} (\textit{i.e.}, CowScout), Allflex Livestock Intelligence Global\footnote{Available at \url{https://www.allflex.global}, October 2023.} (\textit{i.e.}, \textsc{scr}), IceRobotics\footnote{Available at \url{https://www.peacocktechnology.com}, October 2023.} (\textit{i.e.}, IceQube and IceTag), and by Madero Dairy Systems\footnote{Available at \url{https://maderodairysystems.com}, October 2023.} (\textit{i.e.}, Pedometer Plus). The latter devices' technical findings report superior accuracy in detection tasks such as localization compared to manual approaches \cite{Lovarelli2020}.

More complete solutions include the herd robot by DeLaval\footnote{Available at \url{https://corporate.delaval.com}, October 2023.} which has several technical capabilities: (\textit{i}) data management related to pregnant animals and those at the cyst and protracted anestrus risk and abortions, (\textit{ii}) insemination timing prediction, (\textit{iii}) progesterone in milk detection, etc. Similarly, Astronaut device by Lely\footnote{Available at \url{https://www.lely.com}, October 2023.} performs milk analysis (\textit{e.g.}, electrical conductivity). With the body condition score (\textsc{bcs}) camera also by DeLaval \cite{Mullins2019}, cow identification is performed with radio frequency thanks to the transponders placed in the animals. The technology behind this solution involves light coding with infrared ray dots and a corporate \textsc{ai}-based algorithm to translate the image data into \textsc{bcs} information.

Among the most novel diagnostic devices, we must mention the milk checker by \textsc{foss} Analytics\footnote{Available at \url{https://www.fossanalytics.com}, October 2023.}, and mastitis detectors by Dramiski\footnote{Available at \url{https://www.draminski.es}, October 2023.}, and Afimilk\footnote{Available at \url{https://www.afimilk.com}, October 2023.}. The latter solutions rely on physicochemical–biological milk/udder alteration analysis and bio-markers from body fluids (\textit{e.g.}, serum) related to mastitis. All of them with satisfactory performance \cite{Dzermeikaite2023}.

For identification purposes, CattleFaceNet has reported the most competitive results regarding accuracy and processing time \cite{Xu2022}. The technological background of this solution involves a Single Stage Headless (\textsc{ssh}) detection network for alignment and face detection multi-task learning with ArcFace loss function to achieve improved feature selection.

A. S. Ali \textit{et al.} (2022) \cite{Ali2022} presented another relevant technological finding with \textsc{ment-egas}. The latter prototype uses electronic nose technology for estrus detection from odor sampling. The Principal Component Analysis (\textsc{pca}) technique was exploited to discard estrus from pro-estrus and met-estrus.

Regarding the use of \textsc{ml} in the dairy industry, there exist a wide variety of applications (\textit{e.g.}, behavior detection \cite{Riaboff2020}; conception, fertility and birth prediction \cite{furukawa2023analysis}; disease diagnosis \cite{Eckelkamp2019}; physiological condition classification \cite{giannuzzi2023prediction}, etc.) thanks to the combination of data from the animals and the environment \cite{Eckelkamp2019}. However, often, the technologies that derive from \textsc{ai} are costly to implement \cite{Heo2019} or inappropriate for the farming industry (\textit{i.e.}, they produce stress on the animals \cite{Mendes2021}). Note that invasive methods have a detrimental effect on cattle productivity \cite{Mendes2021}, such as milk production. Conversely, a higher comfort of the animals results in a reduced need for care and its associated cost. The direct outcome is the more extended longevity of the animals and their yield \cite{agrawalprecision}.

Thus, there still exists the urge to provide robust automation at low cost to this industry moved by the tremendous economic relevance of farm monitoring towards the welfare of the cattle and in order to enable optimal decision-making of the farmers (\textit{i.e.}, the animal welfare and cost trade-off) \cite{Mendes2021}. 

Unfortunately, manual monitoring is still widespread despite being a laborious and time-consuming approach. In contrast, the data gathered by dairy industry sensors has enabled predictive management and early diagnosis detection compared to traditional monitoring approaches \cite{vidal2023impact}. More in detail, novel automated data collection approaches enable the generation of records at a higher frequency in a less controlled manner \cite{vidal2023impact}. Accelerometers have been previously exploited to study general \cite{Riaboff2020} and lying \cite{Schmeling2021} behavior, health conditions \cite{Haladjian2018}, and rumination patterns \cite{Iqbal2021}.

Mainly, there exist \textsc{ml} solutions to detect lameness, the third most crucial disease\footnoteref{mckinsey} \cite{Taneja2020} and mastitis \cite{Dhoble2019}, birth prediction (also named calving time in the dairy industry) \cite{furukawa2023analysis} using data gathered from sensors in the animals such as activity detectors or the robots like the milking ones, among other applications. In farming monitoring, decision tree models are the most popular approach, followed by artificial neural networks (\textsc{ann}s) to detect diseases and milk quality and production \cite{Slob2021}. Regression algorithms are almost exclusively used as benchmark models \cite{Slob2021}. Note that in the particular case of the dairy industry, high specificity is preferred to reduce the false positive events \cite{DeVries2023}. 

One of the main concerns on the performance of \textsc{ml} solutions in the dairy sector is the quality and lack of training data along with the availability of data streams from several farms gathered in longer time spans, as noted by M. Cockburn (2020) \cite{Cockburn2020}. Notably, integrating the information from different sources must be addressed to take full advantage of the data. In this line, R. García \textit{et al.} (2020) \cite{Garcia2020} stated that mature \textsc{ml} farming solutions are still in development.

\subsection{Discussion of recent \textsc{ml} solutions in the dairy industry}
\label{sec:ml_solutions}

Even though data storage, networking elements, and sensor technologies are already mature solutions, new data analytical models to promote progress are needed in the dairy sector \cite{Norton2019}.

When it comes to the application of \textsc{ai} and \textsc{ml} potential, data fusion has reported significant improvement not only in the dairy industry \cite{Hayes2023,Li2023,Zheng2023} but also in several industrial sectors (\textit{e.g.}, automotive \cite{jaen2023statistical}, construction \cite{Tang2023}, and food industry \cite{Zarezadeh2023}) in which data comes from many different sources and instruments that when combined can provide compositional valuable insights \cite{Zhang2020}. However, incorporated noise in the data must be monitored so as not to affect subsequent analysis. Proper data fusion requires dimensionality reduction and feature selection for each data source using \textsc{pca} vectors and Partial Least Squares regression (\textsc{pls}) variables. Even though data fusion is mainly applied to classification problems, clustering techniques can also be combined to avoid the need for labeled data. In this regard, the most popular clustering technique is the traditional k-means algorithm used as a screening model in the dairy industry \cite{Hayes2023}.

Popular supervised classification methods in the dairy industry (\textit{i.e.}, regression and classification models) include \textsc{pls}-Discriminant Analysis (\textsc{pls-da}) which encompasses dimensional reduction similar to \textsc{pca} but for categorical data \cite{giannuzzi2023prediction,Hayes2023,frizzarin2023estimation}, \textsc{ann}s \cite{giannuzzi2023prediction,Li2023,Zheng2023,frizzarin2023estimation}, k-Nearest Neighbors (k\textsc{nn}) \cite{vidal2023impact,Zheng2023,brouwers2023towards} and tree-based models like Random Forest (\textsc{rf}) \cite{furukawa2023analysis,giannuzzi2023prediction,vidal2023impact,Zheng2023}. Note that \textsc{ann}s stands for non-linear modeling, and its most popular variant in this field is backpropagation (\textsc{bpnn}) \cite{Hayes2023}. All the latter techniques have also reported promising performance in other industries (\textit{e.g.}, automation in productive processes \cite{Schweitzer2023} and predictive maintenance \cite{gupta2023predictive}).

A new trend in \textsc{ml} is eXplainable techniques to understand the black-box models. More in detail, the interpretability needs favor statistical \textsc{ml} (\textit{i.e.}, linear and logistic regression used as a baseline for other \textsc{ml} models like kernel methods as Support Vector Machine - \textsc{svm} and tree-based ones as \textsc{rf}) which is exploited in scenarios were the input data is scarce. Otherwise, non-statistic \textsc{ml} (\textit{e.g.}, \textsc{ann}s, deep neural networks) is applied (\textit{i.e.}, in image processing). Moreover, when applying \textsc{ml} techniques in offline or online mode, reinforcement learning can be considered as online supervised learning since the model reacts to new input data by acquiring new configurations \cite{DeVries2023}.

Among the most recent \textsc{ml}-based solutions for the dairy industry, the work by E. Furukawa \textit{et al.} (2023) \cite{furukawa2023analysis} deserves attention. The authors performed calving prediction using \textsc{ml} models and ruminal temperature (\textsc{rt}) data from 25 pregnant cows. The solution attained \SI{77.8}{\percent} precision in cross-validation. Furthermore, D. Giannuzz \textit{et al.} (2023) \cite{giannuzzi2023prediction} designed a competing blood metabolites (\textit{e.g.}, energy metabolism, liver function, oxidative stress, etc.) prediction solution combining milk Fourier-transform mid-infrared (\textsc{ftir}) spectra and \textsc{ml} models. The data came from multiple farms and 2741 cows of 33 different herds.

Conversely, G. Vidal \textit{et al.} (2023) \cite{vidal2023impact} performed metritis prediction using multiple time windows and several \textsc{ml} models. Input data from an accelerometer in 138 lactating cows comprised activity level (active, highly active) and type (eating, not active, ruminating). Note that the best \textit{F}-measure values were obtained with the tee-based \textsc{rf} algorithm. Moreover, S. P. Brouwer \textit{et al.} (2023) \cite{brouwers2023towards} studied the atypical postural behavior of cows (\textit{i.e.}, lying down and standing up). For that purpose, the authors trained \textsc{ml} models with data gathered from accelerometers placed on the animals, particularly from 48 lactating cows from two dairy breeds. The records were labeled using video observation. The InceptionTime Deep Learning model obtained the best results, reaching more than \SI{70}{\percent} accuracy in a balanced scenario. However, the authors reported that the obtained performance was not yet satisfactory for real applications in farms since human-designed ethograms were not optimal for \textsc{ml}. Similarly, Zheng \textit{et al.} (2023) \cite{Zheng2023} developed a Siamese attention model (Siam-\textsc{am}) with the dual objective of analyzing motion behavior and detecting lameness.

M. Frizzarin \textit{et al.} (2023) \cite{frizzarin2023estimation} analyzed the Body condition score (\textsc{bcs}) (\textit{i.e.}, cow body reserves) using milk mid-infrared spectra (\textsc{mir}) and Neural Networks over the records of 6572 cows in 5 different farms. Q. Li \textit{et al.} (2023) \cite{Li2023} proposed a temporal aggregation network using micromotion features for lameness detection, while Z. Y. Wang \textit{et al.} (2023) \cite{Wang2023} proposed efficient 3D \textsc{cnn} and efficient channel attention (\textsc{eca}) for video processing and information filtering, respectively. The authors focused on activity and behavior analysis (eating, drinking, lying, standing, walking).

Table \ref{tab:articles} summarizes the key aspects of the contributions above for comparison purposes. The numerical results vary notably among applications and methodologies regarding metrics used and values obtained. However, all are above the \SI{70}{\percent} threshold.

\begin{table}[!htbp]
\centering
\footnotesize
\caption{\label{tab:articles} Key aspects of the selected works from the literature.}
\begin{tabular}{p{3.5cm}p{2.5cm}p{2cm}p{2.5cm}p{3.5cm}} 
\toprule
\bf Authorship & \bf Application & \bf Input data & \bf Methodology & \bf Results\\\midrule

E. Furukawa \textit{et al.} (2023) & \multirow{2}{*}{Calving} & \multirow{2}{*}{\textsc{rt}} & \multirow{2}{*}{Supervised \textsc{ml}} & \multirow{2}{*}{\SI{78}{\percent} precision}\\
\cite{furukawa2023analysis}\\\\

D. Giannuzz \textit{et al.} (2023) & \multirow{2}{*}{Blood metabolites} & \multirow{2}{*}{\textsc{ftir} spectra} & \multirow{2}{*}{Supervised \textsc{ml}} & \multirow{2}{*}{\SI{72}{\percent}-\SI{87}{\percent} \textsc{r}$^2$}\\
\cite{giannuzzi2023prediction}\\\\

G. Vidal \textit{et al.} (2023) & \multirow{2}{*}{Metritis} & \multirow{2}{*}{Accelerometers} & Supervised \textsc{ml} & \SI{90}{\percent}-\SI{100}{\percent} accuracy\\
\cite{vidal2023impact} & & & + time windows & \SI{79}{\percent}-\SI{100}{\percent} \textsc{f}-score\\\\\\

S. P. Brouwer \textit{et al.} (2023) & \multirow{2}{*}{Postural behavior} & \multirow{2}{*}{Accelerometers} & Supervised \textsc{ml} & \multirow{2}{*}{\SI{71}{\percent} accuracy}\\
\cite{brouwers2023towards} & & & Deep Learning\\\\

Z. Zheng \textit{et al.} (2023) & Motion behavior & \multirow{2}{*}{Video data} & Supervised \textsc{ml} & \SI{94}{\percent} accuracy\\
\cite{Zheng2023} & Lameness & & Neural Networks & \SI{95}{\percent} accuracy\\\\

M. Frizzarin \textit{et al.} (2023) & \multirow{2}{*}{\textsc{bcs}} & \multirow{2}{*}{\textsc{mir} spectra} & \multirow{2}{*}{Neural Networks} & \multirow{2}{*}{\SI{82}{\percent}-\SI{87}{\percent} correlation}\\
\cite{frizzarin2023estimation}\\\\

Q. Li \textit{et al.} (2023) & \multirow{2}{*}{Lameness} & \multirow{2}{*}{Video data} & \multirow{2}{*}{Neural Networks} & \multirow{2}{*}{\SI{99}{\percent} accuracy}\\
\cite{Li2023}\\\\

Y. Wang \textit{et al.} (2023) & \multirow{2}{*}{Motion behavior} & \multirow{2}{*}{Video data} & \multirow{2}{*}{Neural Networks} & \SI{98}{\percent} precision\\
\cite{Wang2023} & & & & \SI{97}{\percent} recall\\\\

\bottomrule
\end{tabular}
\end{table}

\subsection{Present challenges}
\label{sec:challenges}

The challenges of the existing models applied to the dairy industry are five-fold:

\begin{description}

    \item \textbf{Compatibility and interoperability}. New technologies applied to precision farming nowadays are developed by different providers, causing integration problems. Note that the latter applies not only to physical devices (\textit{e.g.}, \textsc{ai}-driven sensors and robots) but to software platforms (\textit{e.g.}, programs for data analytics). New standards need to be developed in a collaboration framework among vendors.

    \item \textbf{Data privacy \& security}. The vast amount of data gathered from all the devices composing the dairy farming infrastructure contains sensible information regarding business operations, practices, and cattle parameters. Thus, access control, cybersecurity measures, and data encryption algorithms must further be exploited. More in detail, \textsc{i}o\textsc{t} has extensive vulnerabilities derived from the remote control of connected machinery. Therefore, software update processes and a secure network configuration are essential to avoid \textsc{dd}o\textsc{s} attacks or those trying to gather sensitive data.

    \item \textbf{Implementation costs}. The initial investment in precision farming is great in terms of hardware infrastructure and software. To address the latter challenge, the farmers are offered financial incentives to support adopting new technologies in the industry. Moreover, as in other sectors, the continuous technological advancements lead to more affordable solutions when they become mainstream thanks to increased computing and storage capacity of new \textsc{ai}-based models. Thus, the cost is predicted to decrease over time. More in detail, within the competitive economic frames \textsc{ai} solutions will become more accessible.

    \item \textbf{Specialized training}. Specialized skills are needed to ensure the proper integration of \textsc{ai} technologies in the dairy industry (\textit{e.g.}, data analytic training to take advantage of the great potential of \textsc{ml} models). The latter must be addressed with specialized training programs on \textsc{ai} technologies. The automation processes provided by these technologies and the increase in production and lower costs due to their application will allow the inclusion of new actors in the field, such as security professionals and \textsc{ai} experts. This inclusion will represent a challenge of integration between areas of knowledge and sectors that have until now been slightly connected towards a multidisciplinary industrial environment.
    
    \item \textbf{Application of environmental-responsible and sustainable practices}. This can be performed to minimize the environmental impact through reducing gas emissions and resource optimization. Note that \textsc{ai}-driven solutions promote animal welfare (\textit{e.g.}, new milking robots which respect the cows' behavior and rhythms). The latter will undoubtedly contribute to a more ethical and humane farming practice.

\end{description}

\section{Conclusions}
\label{sec:conclusions}

The industry has recently started to acknowledge the great potential of industrial informatics toward productivity enhancement through automation. Regarding the dairy industry, the application of \textsc{ai} techniques in farming is expected to be extensive shortly due to the production improvement mentioned above and expense reduction.

In this line, \textsc{ai} techniques and \textsc{ml} models are appropriate to analyze production and performance in digitized industrial sectors. However, some \textsc{ai} methodologies are costly to implement or inappropriate for the farming industry (\textit{i.e.}, they are considered invasive and detrimental to the cattle). Consequently, a need for robust automation still exists, paying particular attention to animal welfare and cost trade-offs.

This work provides a comprehensive description of industrial challenges where \textsc{ai} can be exploited in the dairy sector, discussing the most recent works in the literature. Summing up, data-related (\textit{e.g.}, missing or unbalanced data, management of data from different farms, etc.) were identified as the most common challenges. Another relevant concern is knowledge extraction and generation from the data collected.

In future work, we plan to analyze the application of genetic engineering and its impact on animals in the long term. The digital twin paradigm in the dairy industry will be studied since no works have been identified. Thus, we plan to improve the technological scope of our research further by incorporating new \textsc{ai}-based technological findings and applications motivated by the rapidly evolving character of the industry (\textit{i.e.}, incorporating the latest data analysis techniques) and paying special attention to feasible transferring methodologies. Of particular interest will be the study of the impact of the 6\textsc{g} paradigm in the sector regarding the dynamic scalability and significant dimensionality challenges. Finally, the environmental impact of cloud computing and wearable sensors on energy consumption will also be discussed. We open the possibility of performing a similar analysis of \textsc{ai}-based solutions in other relevant farming sectors that play a crucial role in the supply chain, such as the food industry.

\section*{Acknowledgements}

This work was partially supported by Xunta de Galicia grants ED481B-2021-118 and ED481B-2022-093.

\bibliographystyle{IEEEtran}
\bibliography{mybibfile}

\end{document}